%% file: main.tex
\documentclass[conference]{IEEEtran}

\usepackage{times}
\usepackage{graphicx}
\graphicspath{{./figures/}}
\usepackage{amsfonts,amsmath, amssymb}
\usepackage{relsize,setspace}  % used by latex(describe( ))
\usepackage{url}               % may be used in bibliography
\usepackage{fancyhdr}
\usepackage{booktabs}
\usepackage{multirow}
\usepackage{wrapfig}
\usepackage[numbers,sort]{natbib}
\usepackage[pdftex,bookmarks]{hyperref}
\hypersetup{colorlinks,linkcolor={blue},citecolor={blue},urlcolor={blue}}
\usepackage{flushend}
\usepackage{cleveref}
\usepackage[textsize=tiny]{todonotes}
\usepackage{caption}
\usepackage{subcaption}
\usepackage{pdfpages}
\usepackage{soul}
\crefname{figure}{Fig.}{Figs.}
\crefname{section}{Section}{Sections}
\crefname{table}{Table}{Tables}
\usepackage[numbers]{natbib}
\usepackage{multicol}
\usepackage{custom_math}
\usepackage{amsthm}
\theoremstyle{definition}

\newcommand{\comment}[1]{}

\begin{document}

% paper title
\title{An Overview of the Burer-Monteiro\\Method for Certifiable Robot Perception}

\author{Alan Papalia$^{1,2}$, Yulun Tian$^{2,3}$, David M. Rosen$^{1}$, Jonathan P. How$^{4}$, and John J. Leonard$^{4}$\\
    % $^{1}$Massachusetts Institute of Technology $^{2}$Northeastern University
    \small{$^{1}$Northeastern University, $^{2}$University of Michigan, $^{3}$University of California, San Diego, $^{4}$Massachusetts Institute of Technology}
}
% \thanks{\include{sections/acknowledgements.tex}}
\thanks{Sample text}

\maketitle

\input{sections/abstract}

% \IEEEpeerreviewmaketitle

\input{sections/intro}

\input{sections/notation.tex}

\input{sections/what_is_bm.tex}

\input{sections/applications.tex}

\input{sections/practical_details.tex}

\input{sections/open_directions.tex}
\input{sections/conclusion.tex}

\input{sections/acknowledgements.tex}

\small
\bibliographystyle{unsrtnat}
\bibliography{references}

\end{document}

%% file: sections/abstract.tex
\begin{abstract}
    This paper presents an overview of the Burer-Monteiro method (BM), a technique that has been applied to solve robot perception problems to certifiable optimality in real-time. BM is often used to solve semidefinite programming relaxations, which can be used to perform global optimization for non-convex perception problems.  Specifically, BM leverages the low-rank structure of typical semidefinite programs to dramatically reduce the computational cost of performing optimization.  This paper discusses BM in certifiable perception, with three main objectives: (i) to consolidate information from the literature into a unified presentation, (ii) to elucidate the role of the linear independence constraint qualification (LICQ), a concept not yet well-covered in certifiable perception literature, and (iii) to share practical considerations that are discussed among practitioners but not thoroughly covered in the literature. Our general aim is to offer a practical primer for applying BM towards certifiable perception.
\end{abstract}
% keywords
\begin{IEEEkeywords}
    Burer-Monteiro,
    Semidefinite Programming,
    Certifiable Perception,
    Riemannian Staircase
\end{IEEEkeywords}

%% file: sections/intro.tex
\section{Introduction}

% 1) perception algorithms need to be (a) reliable and (b) real-time
% 2) certifiably correct algorithms provide formalism for reliability
% 3) in many cases, BM provides a way to efficiently construct
% certifiably correct algorithms
Robotic perception is increasingly entering the world in a range of
applications, from augmented reality to autonomous vehicles. These algorithms
need to be both (i) real-time operable and (ii) safe and trustworthy. However,
these aims are often at odds with one another, as providing rigorous
guarantees on reliability often comes at substantial computational expense.

Certifiably correct algorithms -- algorithms with formal guarantees on the
\textit{correctness} (i.e., global optimality) of their solutions -- provide one
means of providing reliability. However, many certifiably correct algorithms are
computationally infeasible for real-time operation. This paper discusses the
Burer-Monteiro method (BM), a technique that has been successfully applied to
robotic perception problems to construct \textit{computationally
      efficient} certifiably correct algorithms.

BM is a framework for solving large-scale semidefinite
programming (SDP) relaxations; these SDP relaxations often arise as convex
relaxations of (non-convex) perception problems. SDP relaxations are
useful, as they provide strong theoretical frameworks for analyzing the
optimality of solutions. However, despite being convex, standard SDP solvers
do not scale well to large problems. BM provides a path to efficiently
solve these SDP relaxations by exploiting the fact that solutions
to these relaxations are often low-rank (as is typically the case in perception
problems).

This paper is motivated by the growing amount of research in certifiably correct
perception and the complex landscape of algorithms that have been developed in
this area. In particular, this paper aims to convey \textit{when} and
\textit{why} BM is suitable for certifiably correct
algorithms, clarifying common points of confusion and providing a roadmap for
future research.

This paper does not convey new research results, nor does it extensively cover
theoretical details on BM or SDP relaxations. Instead, this paper seeks to
lower the barrier to entry into this field by:
\begin{itemize}
      \item \emph{unifying knowledge} from the literature into a single, introductory
            discussion of BM and certifiable perception,
      \item highlighting the often unstated importance of the
            \emph{linear independence constraint qualification} (LICQ)
            to a greater extent than currently found in the robotics literature, and
      % \item highlighting the often unstated importance of the
      %       \emph{linear independence constraint qualification} (LICQ), and
      \item sharing \emph{practical considerations} for applying BM to real-world problems.
\end{itemize}

% \begin{itemize}
%       \item theoretical guarantees provided by SDP relaxations,
%       \item mathematical conditions under which BM can be efficiently
%             applied,
%       \item how to efficiently perform certification with BM,
%       \item computational algorithms for solving BM problems,
%       \item applications where BM has been successfully
%             applied,
%       \item practical considerations for applying BM to real-world
%             problems, and
%       \item future research directions to advance the state of the art in
%             certifiably correct perception.
% \end{itemize}

%% file: sections/notation.tex
\section{Notation}
\label{sec:notation}

We briefly introduce the notation used in this paper. We use lowercase letters
to denote vectors, e.g., $x$, and uppercase letters to denote matrices, e.g.,
$X$. We use $\innerProd{\cdot}{\cdot}$ to denote the inner product, where the
matrix inner product is defined as $\innerProd{A}{B} = \Tr(A^\top B)$ and which
simplifies to the standard Euclidean inner product for vectors $\innerProd{x}{y}
= x^\top y$. $\Sym^n$ denotes the space of $n \times n$ real, symmetric
matrices. The relationship $A \succeq 0$ means that $A$ is positive
semidefinite.

%% file: sections/what_is_bm.tex
\section{The Burer-Monteiro Method (BM)}

In this section we demonstrate how the Burer-Monteiro method (BM) is directly
derived from standard \textit{maximum a posteriori} estimation problems that
arise in perception, and how BM enables efficient global optimization. Most of
the material in this section is discussed across various papers in the
literature (e.g.,
\cite{burer2003nonlinear,rosen2021scalable,tian2021distributed,papalia24tro});
this section concisely gathers this knowledge under a single framework.
Additionally, we contain a small discussion (\cref{sec:licq}) on the relevance
of the \textit{linear independence constraint qualification} (LICQ).

\input{sections/general_workflow}

\subsection{Local Solvers Compatible with BM}
\label{sec:local_solvers}
\input{sections/local_solvers}

\input{sections/certification}

\input{sections/licq}

%% file: sections/general_workflow.tex
\subsection{From Perception to Certifiable Optimization and the BM}

We begin by showing how a quadratically constrained quadratic program (QCQP) can
be relaxed to a (convex) semidefinite program (SDP) via a
standard technique (Shor's relaxation \cite{shor1987quadratic}). We
discuss the theoretical benefits conveyed by this relaxation and the
computational challenges it introduces. Importantly, we discuss how the SDP
relaxation provides a useful \textit{sufficient} condition for the global
optimality of the original QCQP.

We then introduce BM, a (non-convex) low-dimensional factorization of the SDP,
to address the computational challenges of SDPs. We show how BM is
naturally derived from the SDP relaxation and the benefits it provides to
solving SDPs. We discuss theoretical properties of BM and how specific
conditions can be evaluated to see if a locally optimal BM solution corresponds
to a globally optimal solution for the corresponding SDP.

\textbf{QCQPs in perception.}
Our paper starts by assuming that a perception problem is formulated as a
quadratically constrained quadratic program (QCQP), i.e., a problem with
both quadratic objective and constraint functions. Many applications in robot perception
satisfy this criterion (see \cref{sec:applications}).
 In our paper, this is a
non-convex QCQP (otherwise, the machinery discussed here is not necessary).
For conciseness, we will consider QCQPs with only equality constraints, though
the methods we discuss readily extend to inequality constraints
\cite{rosen2021scalable}. We will consider QCQPs posed as follows:
\begin{equation}
    \label{eq:qcqp}
    \begin{aligned}
        \min_{X \in \R^{n \times k}} & \quad \innerProd{Q}{X X^\top}                                  \\
        \text{s.t.}                  & \quad \innerProd{A_i}{X X^\top} = b_i, \quad i = 1, \ldots, m.
    \end{aligned}
\end{equation}
$Q \in \Sym^{n}$
is a matrix that encodes quadratic costs, often
called a data matrix in perception, and $A_i \in \Sym^{n}$ are matrices
that encode quadratic constraints (which may also depend on data). The $b_i \in \R$ are constants that encode the
right-hand side of the constraints. We note that the expression $\innerProd{A}{X
        X^\top}$ is equivalent to the more familiar quadratic form $\Tr(X^\top A X)$,
which simplifies to $x^\top A x$ in the case of vectors.

Admittedly, much of the art of certifiable perception is in finding the right
QCQP formulation for a given problem. We will not delve into this in this
paper, but we point out that
quadratic constraints are common in perception. E.g., the orthogonality of
rotation matrices $(R^\top R = I)$ or the unit norm of quaternions $(q^\top q = 1)$.

\textbf{From QCQP to SDP (Shor's relaxation).}
Given a QCQP of the form above, we can follow a well-known procedure to relax
it to a semidefinite program (SDP). This relaxation is known as Shor's
relaxation \cite{shor1987quadratic}.
The idea is to first introduce a variable
substitution $Z = X X^\top$ and rewrite the QCQP as an equivalent (non-convex)
SDP:
\begin{equation}
    \label{eq:rank-constrained-sdp}
    \begin{aligned}
        \min_{Z \in \Sym^n} & \quad \innerProd{Q}{Z}                                 \\
        \text{s.t.}         & \quad \innerProd{A_i}{Z} = b_i, \quad i = 1, \ldots, m \\
                            & \quad Z \succeq 0,
        \\
                            & \quad \rank(Z) \leq k
    \end{aligned}
\end{equation}
where the implicit properties of the outer product $X X^\top$ are explicitly encoded
as constraints on $Z$ in the form of symmetry $Z \in \Sym^n$, positive
semidefiniteness $Z \succeq 0$, and rank $\rank(Z) \leq k$.

The sole source of non-convexity in \eqref{eq:rank-constrained-sdp} is the rank
constraint
\cite{vandenberghe1996semidefinite}. By dropping the rank constraint, we obtain
a convex SDP relaxation of the original QCQP.
\begin{equation}
    \label{eq:SDP}
    \begin{aligned}
        \min_{Z \in \Sym^n} & \quad \innerProd{Q}{Z}                                 \\
        \text{s.t.}         & \quad \innerProd{A_i}{Z} = b_i, \quad i = 1, \ldots, m \\
                            & \quad Z \succeq 0                                      \\
    \end{aligned}
\end{equation}
This relaxation is useful because it provides a \textit{sufficient} condition
for the global optimality of the original QCQP. Because the SDP is a relaxation
of the QCQP, any solution to the SDP must lower-bound the optimal value of the
QCQP.  As a result, if we can find a point $X^*$ for the QCQP \eqref{eq:qcqp}
such that $Z^* = X^* X^{*\top}$ solves the SDP \eqref{eq:SDP}, then we know that
(i) the optimal value of the QCQP and the SDP are the same and (ii) $X^*$ is a
global optimum of the QCQP. This simple idea has powerful implications. Namely,
this provides a path to efficient \textit{global optimization} of a non-convex
problem. Rather than solving the (non-convex) QCQP directly, we can solve the
(convex) SDP and try to extract a QCQP solution $X^*$ from the SDP solution
$Z^*$ via e.g., singular value decomposition. However, there is
a catch: the SDP relaxation is often too computationally expensive to solve
with standard solvers. This computational challenge is where BM becomes
useful.

\textbf{The Burer-Monteiro method.}
BM is a technique to reduce the computational complexity of solving SDPs which
have a low-rank structure, i.e., SDPs where the solution has low rank relative
to the problem size: $\rank Z^* \ll n$.
Fortunately, the SDP relaxations that arise from QCQPs in perception often have
such low-rank structure (this can often be proven for certain noise regimes
e.g., \cite{rosen2019se,cifuentes2022local}).

BM simply follows by introducing a factorization $Z = Y Y^\top, Y \in \R^{n
        \times r}$ into the SDP relaxation. This factorization introduces two
\textit{implicit} constraints on $Z = Y Y^\top$: (i) positive semidefiniteness
$Z \succeq 0$ and (ii) low-rank $\rank(Z) \leq r$. The resulting BM problem is,
\begin{equation}
    \label{eq:bm-problem}
    \begin{aligned}
        \min_{Y \in \R^{n \times r}} & \quad \innerProd{Q}{Y Y^\top}                                 \\
        \text{s.t.}                  & \quad \innerProd{A_i}{Y Y^\top} = b_i, \quad i = 1, \ldots, m
    \end{aligned}
\end{equation}
With $r \ll n$, the BM problem has a
much lower-dimensional state space than the original SDP. This reduces the
number of computational operations required to solve the problem, making it more
tractable.  Modern BM solvers can operate on problems with $n$ in the tens of
thousands on standard laptops in seconds. This is in stark contrast to generic
SDP solvers, which on similar machines can struggle with problems of size $n$ in
the hundreds \cite{majumdar2020recent}.

Formulation \eqref{eq:bm-problem} may look familiar; this is effectively the same
factorization we used to relax the QCQP to the SDP in the first place.
Specifically, when the BM variable $Y$ has the same number of columns as $X$ in
the original QCQP (i.e., $r = k$), the BM problem is equivalent to the original
QCQP. In general, the BM problem is also a non-convex QCQP.
This may seem like a circular way to arrive at the original problem, but this
circuitous route provides us two key viewpoints.

% a hierarchy of relaxations of the QCQP
First, we can see that by increasing the rank $r$ of the BM problem, we can view
BM as providing a \textit{hierarchy of relaxations} of the original problem.
Intuitively, increasing $r$ increases
the free dimensions that an optimizer may access, allowing for new descent
directions that can avoid what are local minima in lower-rank relaxations of
the BM problem.

% a way to certify global optimality
The second advantage is the capacity for \textit{efficient global optimization}.
This stems from connection to the SDP relaxation.
Formally, the following relationship holds:
\begin{equation}
    \label{eq:optimality-relationship}
    f^*_\text{SDP} \leq f^*_\text{BM} \leq f^*_\text{QCQP}.
\end{equation}
Where $f^*_\text{SDP}$, $f^*_\text{BM}$, and $f^*_\text{QCQP}$ are the optimal
values of the SDP, BM, and QCQP problems, respectively and the BM problem is
not lower-dimensional than the original QCQP (i.e., $r \geq k$).  This
relationship provides a sufficient condition to \textit{certify} the global optimality of BM and
QCQP solutions in the case when the inequalities are tight (i.e.,
$f^*_\text{SDP} = f^*_\text{BM} =
    f^*_\text{QCQP}$).

With this in mind, we can solve a series of BM problems with increasing rank $r$
until a solution $Y^*$ is found that is also a low-rank factorization of the SDP
solution $Z^* = Y^* Y^{*\top}$, an approach known as the \textit{Riemannian
Staircase} \cite{boumal2015riemannian} when the BM problems are solved using
Riemannian optimization. As we have a solution to the SDP, we have
$f_\text{SDP}$. Because $Y^*$ attains the same objective value as $Z^*$, from
the relationship in \eqref{eq:optimality-relationship}, we also know we have a
globally optimal solution to the BM problem with objective value $f_\text{BM} =
f_\text{SDP}$. Finally, if $Y^*$ has the same rank as the dimension of the QCQP
\eqref{eq:qcqp} (i.e.,
$\rank(Y^*) = k$), then we have a globally optimal solution to the original QCQP
with objective value $f_\text{QCQP} = f_\text{BM} = f_\text{SDP}$.\footnote{
    We can always obtain a certifiably optimal BM solution $Y^*$. However, it
    is possible that the SDP relaxation is not tight (i.e., $f^*_\text{SDP} <
        f^*_\text{QCQP}$ and $\rank(Y^*) > k$); in this case, QCQP solutions
    cannot be certified but \eqref{eq:optimality-relationship} can be used to
    bound the QCQP solution's suboptimality. Additionally, an
    \textit{approximate} solution to the QCQP can be extracted from the BM
    solution $Y^*$ via singular value decomposition,
    subsequently projected to the QCQP's feasible set,
    and used as a starting point for further
    optimization. This has been used to great effect in practice
    (e.g., \cite{papalia24tro}).
}

To take advantage of this theoretical framework, we need two algorithmic tools:
(i) a way to perform optimization on BM problems and (ii) a way to
\textit{certify} whether a BM solution maps to an SDP solution. As we are
concerned with runtime, each of these items must be computationally
efficient. We discuss these two items in the following sections.

% The first item (optimization) is straightforward; we can apply a range of
% local-optimization methods to the BM problem. We discuss different solvers in
% \cref{sec:local_solvers}.

% The second item (certification) is more theoretically involved. We need to
% \textit{efficiently} show that a BM solution $Y^*$ maps to a solution to the
% SDP. We discuss certification in depth in \cref{sec:certification}.

%% file: sections/local_solvers.tex
%!TEX root = ../main.tex

In this subsection, we review optimization algorithms that are suitable for
identifying local solutions to the intermediate problems \eqref{eq:bm-problem}
introduced by the BM hierarchy. These solvers are not limited to BM -- they
can be applied to a wide range of optimization problems beyond the specific
structure of \eqref{eq:bm-problem}. However, the choice of solver has a major
impact on the efficiency and reliability of the overall BM approach. We
highlight the most relevant solvers below.

One approach is to view and (locally) solve \eqref{eq:bm-problem} as an instance of a generic nonlinear program \cite{burer2003nonlinear}.
We refer to this as an \emph{extrinsic} approach, because it
enforces the search space $\mathcal{M} \triangleq \{Y \in \mathbb{R}^{n \times r}: \innerProd{A_i}{Y Y^\top} = b_i, i = 1, \ldots, m\}$ using explicit constraints in the ambient Euclidean space $\mathcal{E} \triangleq \mathbb{R}^{n \times r}$.
The original work of Burer and Monteiro \cite{burer2003nonlinear} uses an extrinsic method to locally solve \eqref{eq:bm-problem}, which enforces the search space constraints by optimizing over the augmented Lagrangian function.
Recent work \cite{rosen2021scalable} extends BM to generic low-rank SDPs with inequality constraints and also adopts an extrinsic local solver implemented in IPOPT \cite{wachter2006implementation}.
% \yulunnote{Move extrinsic solvers here.}
% \alannote{Comments from Alex Amice:
% From what I know, the main method is LBFGS or some variant of it.... other people have taken more of a first-order approach based on douglas-rachford splitting (ADMM)
% The main reason to use a second-order, quasi newton methods if i recall correctly is that the BM formulation has spurious first order stationary points (gradient vanishes), but second-order stationary points are very uncommon (and can be provably excluded in many cases)
% }

\textbf{Intrinsic (Riemannian) solvers.}
In contrast to the extrinsic local solvers above,
more recent works further exploit the geometric structure of \eqref{eq:bm-problem} through an \emph{intrinsic} perspective.
In practice, for many robot perception applications, $\mathcal{M}$ turns out to be ``standard'' matrix manifolds whose geometries are well studied (e.g., a Stiefel manifold \cite{rosen2019se,briales2017cartan,dellaert2020shonan}).
% The theoretical underpinning is that under LICQ, the search space $\mathcal{M}$ (under equality constraints) is a Riemannian submanifold of $\mathcal{E}$ (e.g., see \cite[Proposition~1.2]{boumal2020deterministic}).
% Furthermore, for many robot perception applications, $\mathcal{M}$ turns out to be ``standard'' matrix manifolds whose geometries are well studied (e.g., a Stiefel manifold \cite{rosen2019se,briales2017cartan,dellaert2020shonan}).
As such, existing theories and implementations of \emph{Riemannian optimization}
\cite{absil2008optimization,boumal2023introduction} directly apply. These
approaches solve \eqref{eq:bm-problem} by operating on the manifold
intrinsically.
The intrinsic approach is favorable because the corresponding Riemannian optimization problem is \emph{unconstrained}.
This enables the use of unconstrained optimization algorithms (generalized to operate on manifolds) that by design
produce a sequence of \emph{feasible} iterates,
enjoy convergence guarantees similar to those of extrinsic solvers,
and have empirically shown to be substantially more efficient than
extrinsic solvers in perception applications, e.g., for pose graph optimization
\cite{rosen2021scalable}.
% \yulunnote{Do you know more refs to support the last
% claim? -- I don't, but I believe it to be generally true (having tried implementing things in more general solvers myself)}

% In the following, we review state-of-the-art intrinsic and extrinsic local solvers, and discuss the trade-offs between these approaches for practitioners in robot perception.
% Under LICQ, it can be shown that the search space in \eqref{eq:bm-problem} forms a smooth \emph{manifold} embedded in the Euclidean space.
% \paragraph{Intrinsic (Riemannian) solvers}

For typical instances of \eqref{eq:bm-problem} arising from robot perception
applications, \emph{second-order} (Newton-type) solvers combined with
\emph{globalization} strategies (e.g., trust-region) have proven particularly
effective. The second-order property of the solver helps to evade spurious
first-order critical points (where the gradient vanishes) and achieves a fast
(superlinear) local convergence rate.
The globalization strategy further prevents possible divergence (which is possible with the vanilla Newton's method),
and ensures optimization converges from any initial guess.
% Second-order solvers ensure fast local convergence (similar to Newton's method), and globalization strategies ensures convergence from any initial guess (otherwise, methods such as the vanilla Newton's method could diverge).
% We discuss two prominent examples.
A prominent example that follows this design principle is the Riemannian
trust-region (RTR) algorithm \cite{absil2007trust} used by many state-of-the-art
certifiable methods \cite{rosen2019se,briales2017cartan,papalia24tro}.
At every iteration, RTR approximately minimizes a local second-order model of \eqref{eq:bm-problem} under a trust-region constraint, which limits the magnitude of the computed update.
The size of the trust region is adjusted dynamically, so that it acts as a safeguard when the quality of the model function is poor, but still does not interfere with the fast local convergence of typical second-order optimization.
The model minimization is typically done iteratively using the truncated conjugate gradient (tCG) method (e.g., see \cite[Section~6.3]{boumal2023introduction} for details).
Closely related to RTR is the Riemannian Levenberg-Marquardt (LM) method \cite[Section~8.4.2]{absil2008optimization}.
LM uses the same local quadratic model of the objective as the Gauss-Newton method.
Instead of changing the trust-region size as in RTR, LM dynamically adjusts a regularization term that is added to the model function, which can be interpreted as the \emph{Lagrangian form} of a trust-region constraint and plays a similar role of discouraging large updates.
% Together with the Gauss-Newton method, LM is the workhorse underlying today's high-performance nonlinear least squares libraries such as GTSAM \cite{gtsam}, Ceres Solver \cite{ceres-solver}, and g2o \cite{g2o}.
Among recent certifiable perception methods, the rotation averaging method by \citet{dellaert2020shonan} uses LM as implemented in GTSAM \cite{gtsam} to solve BM problems defined on rotation groups with increasing dimensions.

% \paragraph{Extrinsic solvers}
% In their initial paper \cite{burer2003nonlinear}, Burer and Monteiro use an extrinsic method to locally solve \eqref{eq:bm-problem}, which enforces the search space constraints by optimizing over the augmented Lagrangian function.
% \yulunnote{More inputs needed.}

% \paragraph{Intrinsic vs. extrinsic solvers}
% \paragraph{Extension to distributed/parallel computing}
\textbf{Extensions to distributed/parallel computing.}
To extend the Riemannian Staircase approach
\cite{boumal2015riemannian,rosen2019se} to the distributed regime,
\citet{tian2021distributed} developed Riemannian Block Coordinate Descent (RBCD)
as a distributed local optimization method that leverages the
product manifold structure that naturally arises in many robot perception
tasks.
%  natural
% partitioning of the search space.
 Additional enhancements to this idea,
including an extension to operate under asynchronous communication, are proposed
in \cite{tian2020asynchronous,peng2023block}.  In general, many other
distributed optimization algorithms are theoretically compatible with BM if they
can identify first-order critical points of \eqref{eq:bm-problem}.  Examples
include methods based on distributed Riemannian gradient descent
\cite{tron2009distributed,tron2014distributed,asgharivaskasi2024riemannian}.
Recent works have also extended the alternating direction method of multipliers
(ADMM) to solve distributed optimization over factor graphs
\cite{xu2022d,banninger2023cross,mcgann2023asynchronous}, although formal
convergence guarantees remain to be explored under Riemannian manifold
constraints.  \citet{fan2023majorization} developed an extrinsic method based on
accelerated majorization minimization for distributed pose graph optimization.

%% file: sections/certification.tex
\subsection{Certification}
\label{sec:certification}

\def\S{S_{\lambda}}
\newcommand{\cdash}{{\textemdash}}
\begin{table}[!tbp]
    \centering
    \begin{tabular}{l l l}
                                  & Burer-Monteiro                        & SDP                            \\
        \midrule
        \text{Stationarity}       & $\S Y = \mathbf{0}$                   & $\S = Q + \sum \lambda_i A_i$  \\
        \text{Complementarity}    & \cdash                                & $\S Z = \mathbf{0}$            \\
        \text{Primal feasibility} & $\langle A_i, Y Y^\top \rangle = b_i$ & $\langle A_i, Z \rangle = b_i$ \\
                                  &                                       & $Z \succeq 0$                  \\
        \text{Dual feasibility}   & \cdash                                & $\S \succeq 0$                 \\
    \end{tabular}
    \caption{Optimality conditions for the BM and SDP problems, where $\S
            \triangleq Q + \sum \lambda_i A_i$. A ``\cdash''~in an entry
        indicates that the condition is not relevant to the theory presented
        in this paper.  The BM conditions are necessary for first-order
        optimality, while the SDP conditions are necessary and sufficient
        for global optimality. All conditions can be derived from the
        first-order optimality conditions of the respective problem
        \cite{rosen2021scalable}. When all SDP conditions are satisfied,
        $Z$ and $\S$ are a primal-dual pair of solutions to the SDP relaxation.}
    \label{tab:optimality-conditions}
    \vspace*{-5mm}
\end{table}

As previously mentioned, a key capability is certifying whether
a BM solution $Y^*$ is a low-rank factor for a solution $Z^* = Y^*(Y^*)^\top$ of
the original SDP. If so, the BM solution is guaranteed to be globally
optimal  for the BM problem $f_{\text{BM}} =
    f_{\text{SDP}}$), and we now have a certified lower bound on the attainable cost
of the original QCQP ($f_{\text{BM}} \leq f_{\text{QCQP}}$). This lower bound
allows for certification of a QCQP solution when the SDP relaxation is tight
($f_{\text{QCQP}} = f_\text{BM} = f_\text{SDP}$),
which is often the case in practice.

The naive approach to certifying a BM solution is to generate the corresponding
SDP solution $Z^* = Y^* (Y^*)^\top$ and check whether $Z^*$ satisfies the
Karush-Kuhn-Tucker (KKT)
conditions of the SDP relaxation, which are necessary and sufficient for
optimality of the SDP relaxation \cite{boyd2004convex}. However,
$Z^*$ would be generically dense, and thus incur substantial computational
overhead. We instead describe a separate approach (described in
\cite{rosen2021scalable}) that leverages the
low-dimensional BM factorization $Y^*$.

As described in \cref{tab:optimality-conditions}, comparison of the KKT
conditions of the BM and SDP problems reveals that a first-order stationary
point of the BM problem $Y^*$ can generate a candidate solution to the SDP
problem $Z^* = Y^* (Y^*)^\top$, which automatically satisfies all SDP optimality
conditions \textit{except} for the dual feasibility condition $\S \succeq 0$. As
a result, a BM solution can be certified as globally optimal by checking
positive semidefiniteness of the \textit{certificate matrix},\footnote{Note that when
this certificate matrix is positive semidefinite (and thus all the SDP
optimality conditions in \cref{tab:optimality-conditions} are satisfied),
the certificate matrix is also a solution the SDP dual problem. As a result,
the certification scheme described here can also be viewed as searching for a
dual solution.}
\begin{equation}
    \S = Q + \sum_{i=1}^m \lambda_i A_i,
    \label{eq:dual_certificate}
\end{equation}
where $Q$ is the data matrix describing the original QCQP, and $\lambda_i$ and
$A_i$ are the Lagrange multipliers and constraint matrices. In practice,
the fastest and most reliable way to evaluate $\S \succeq 0$ is to compute
a Cholesky factorization of $\S + \epsilon I$ for a small $\epsilon > 0$, which
will fail if $\S$ is not positive semidefinite.\footnote{
    If $\S$ is not positive semidefinite, then a negative eigenpair of $\S$ can
    be used to construct a second-order descent direction to
    kickstart optimization at the next level ($r+1$) of the BM hierarchy. See
    e.g., \cite{boumal2015riemannian} for more details.
}

To actually compute the Lagrange multipliers $\lambda$ at
a candidate solution $Y^*$, one can use the BM stationarity condition from
\cref{tab:optimality-conditions} which is equivalent to,
\begin{equation}
    \label{eq:stationarity}
    \sum_{i=1}^{m} (A_i  Y^*) \lambda_i = -(Q Y^*),
\end{equation}
and solve a linear system for $\lambda$. Given $\lambda$, the certificate
matrix $\S$ can be computed and $\S \succeq 0$ can be checked.

%% file: sections/licq.tex
\subsection{The Role of the LICQ in BM}
\label{sec:licq}

The
\emph{linear independence constraint qualification} (LICQ)
 is a standard constraint qualification that plays a pivotal role in the
success of the BM framework and is prevalent in many robot perception
applications (\cref{sec:applications}). This subsection briefly
discusses this topic for interested practitioners.

% \alannote{We may want to relocate this discussion of LICQ}
The LICQ is satisfied if the gradients of the constraints
are linearly independent. In the context of BM,
this means that
$\{\nabla
    \innerProd{A_i}{X X^\top} \mid i = 1, \ldots, m \}$ is a linearly independent set.

% We will assume that the LICQ is satisfied at all feasible points of the
% QCQP, which is a stronger assumption than necessary for most of the results we
% discuss but is often found to be true in existing applications.
% \cref{sec:licq} discusses the roles played by LICQ in the BM framework.

\textbf{Local optimization (\cref{sec:local_solvers}).} There are two important
aspects of the LICQ with respect to local optimization: (i) the LICQ is closely
connected to the use of Riemannian optimization and (ii) the LICQ is
tightly connected to the convergence of many local optimization algorithms.

Regarding Riemannian optimization, if the LICQ is satisfied globally (i.e., at
all feasible points) for the BM problem \eqref{eq:bm-problem}, then the search
space $\mathcal{M}$ of \eqref{eq:bm-problem} forms a smooth manifold \cite[Ch.
7]{boumal2023introduction}.  However, it is important to note that the LICQ
alone is not enough for the practical success of Riemannian optimization, and
additional information on the knowledge of $\mathcal{M}$ is needed to have
efficient numerical implementations (more specifically, to implement the
\emph{retraction} operators \cite[Ch. 3]{boumal2014manopt} within Riemannian
optimization).

Regarding the convergence of local solvers, the LICQ is a key ingredient in
establishing efficient convergence for many general-purpose optimization
algorithms (e.g., interior-point methods \cite[Ch. 19.8]{nocedal2006book}). The
LICQ (along with the second order sufficiency condition) is necessary to
guarantee nonsingularity of the primal-dual KKT system matrix, which is key in
establishing superlinear convergence of second-order and Newton-type methods.
This dependence on the LICQ highlights the challenges of efficiently solving the
BM problem in its extrinsic form when the LICQ is not satisfied. However, this
does not preclude the use of methods which do not rely on the LICQ (e.g.,
penalty methods), but may not be as efficient as those that do rely on the LICQ.

\textbf{Certification (\cref{sec:certification}).}
The LICQ is key to performing efficient certification. This is because it is the
weakest condition that is necessary and sufficient for the existence of
\textit{unique} Lagrange multipliers \cite{wachsmuth13licq}.  Recall that the
certificate matrix $\S$ in \eqref{eq:dual_certificate} depends on the Lagrange
multipliers $\lambda$, which, at a candidate solution $Y^*$, are determined by
the stationarity condition \eqref{eq:stationarity}.
However, this linear system only
admits a unique solution $\lambda^*$
 if the LICQ holds \cite{wachsmuth13licq}.

If the LICQ is satisfied, certification is done by
solving the linear system \eqref{eq:stationarity} for $\lambda^*$, then
forming $\S$, and finally evaluating
positive semidefiniteness $\S \succeq 0$. If the LICQ is not satisfied, then it is
possible for there to exist many different $\lambda^*$ that satisfy
\eqref{eq:stationarity}, but only \emph{some} of which may correspond to a positive
semidefinite certificate matrix $\S \succeq 0$.
Without the LICQ, certification
is equivalent to finding an intersection of the affine space defined by
\eqref{eq:stationarity} and the positive semidefinite cone, a
semidefinite feasibility problem in its own right and generally as
expensive to solve as the original SDP \eqref{eq:SDP}.

%% file: sections/applications.tex
\section{Applications}
\label{sec:applications}

\comment{
\begin{itemize}
    \item Rotation/Pose Sync
          \begin{itemize}
              \item SE-Sync \cite{rosen2019se}
              \item Cartan Sync \cite{briales2017cartan}
              \item CPL-SLAM \cite{fan2020cpl}
          \end{itemize}
    \item Landmark SLAM \cite{holmes2023efficient}
    \item Range
          \begin{itemize}
              \item CORA \cite{papalia24tro}
              \item Riemannian Elevator \cite{halsted2022riemannian}
          \end{itemize}
    \item Sensor network localization (from Diego's papers)
    \item Matrix sensing (from Diego's papers)
    \item Essential matrix estimation \cite{karimian2023essential}
\end{itemize}
}

To date, BM has been applied to a variety of perception tasks, including
rotation and pose synchronization \cite{rosen2019se,
    briales2017cartan,fan2020cpl,dellaert2020shonan}, landmark-based SLAM
\cite{holmes2023efficient}, range-aided SLAM \cite{papalia24tro},
sensor network localization \cite{halsted2022riemannian}, essential matrix
estimation for structure from motion \cite{karimian2023essential},
and semantic segmentation via Markov random fields \cite{hu2019accelerated}.
In this section we discuss commonalities across existing BM applications
to try to understand the where and why of BM's success in perception.

Of interest is that all of these applications leveraged well-studied Riemannian
manifolds (e.g., the Stiefel manifold or the unit sphere) in formulating their
problems and solved them in their intrinsic forms via Riemannian optimization.
The only instance in the perception literature that we are aware of which used a
purely extrinsic solver for BM is \cite{rosen2021scalable}, which was done to
compare the performance of extrinsic and intrinsic solvers in the context of
pose-graph optimization. Additionally, \citet{karimian2023essential} posed
essential matrix estimation as optimization over the Stiefel manifold with an
additional constraint (to represent epipolar geometry), though still used
intrinsic descriptions of the problem.

Considering this preference, it is natural to ask \textit{why have no perception
    problems used an extrinsic formulation for local optimization?} This is interesting, as
    arriving at an extrinsic BM formulation requires less work; intrinsic
formulations require identifying manifold structure within the BM extrinsic
formulation \eqref{eq:bm-problem} and defining additional manifold notions (e.g., a retraction
operator).  There are several explanations for this preference towards
intrinsic formulations despite the additional work required. We posit that
this preference is due to a combination of: common problem structure, bias
towards successful formulations, and the availability of optimization software.

\textbf{Geometric structure.} Perception problems typically possess smooth,
geometric structure that is naturally expressed as well-studied manifolds. For
example, orthogonality and unit-norm constraints appear throughout perception --
correspondingly the Stiefel manifold
is ubiquitous in certifiable perception. In fact, all existing
works using BM for certifiable perception can be expressed as optimization
over the Stiefel manifold and Euclidean space.\footnote{
    Note that the unit-sphere is a special case of the Stiefel
    manifold.
}
\citet{karimian2023essential} are a notable partial exception, as they
derive a custom manifold by adding an
additional explicit constraint to the Stiefel manifold.

\textbf{Bias towards successful formulations.} There are often many different
QCQP formulations of the same problem, which typically differ in the tightness
of their SDP relaxations.  While understanding the relationship between a QCQP
and the tightness of its SDP relaxation is an open area of research (e.g.,
\cite{cifuentes2022local}), we have empirically found that (a) many possible
formulations of perception problems are not tight, and (b) the Stiefel manifold
often leads to tight SDP relaxations.  As previously noted, all works to date
have used some formulation that can be related to (special cases of) the Stiefel
manifold. Therefore, this apparent bias towards intrinsic formulations may be
``natural selection'' appearing due to ideal properties of the Stiefel manifold.

\textbf{Available optimizers.}
Existing manifold optimization libraries are relatively mature, allowing for
straightforward evaluation of intrinsic formulations without requiring the user
to implement their own optimizer. In contrast, the apparent lack of a ``standard''
extrinsic solver for BM represents a barrier to evaluating more general
extrinsic formulations in practice. We review available solvers in
\cref{sec:practical_details}.

Additionally, as noted in \cref{sec:local_solvers}, Riemannian
optimization conveys substantial computational benefits over
extrinsic optimization. The combined advantages in reliability and efficiency
of Riemannian solvers further incentivize the additional effort required to
formulate problems intrinsically.

%% file: sections/practical_details.tex
\label{sec:practical_details}

% There are a number of important considerations that aid practitioners in applying
% BM to problems in perception.

% We briefly reiterate some theoretical details that affect performance:
% problem formulation (see \textit{geometric structure} in \cref{sec:applications}),
% certification methodology (see \cref{sec:certification}),
% and optimization algorithm (see \cref{sec:local_solvers}).

There are several practically oriented considerations in applying BM to
certifiable perception problems. We base these considerations on our own
experiences as practitioners. We focus on numerical conditioning, sparsity, and
existing solvers. The first two points (conditioning and sparsity) are
particularly relevant for problems such as SLAM, which often manifest as
large-scale estimation over sparse graphs. The final point (existing solvers) is
useful for all practitioners, as there are many flexible and performant
optimization libraries available.

\textbf{Numerical (pre)conditioning for RTR.}
Many perception problems lead to large and ill-conditioned optimization
problems, which presents a substantial challenge for many optimization
algorithms.
In the following, we focus on RTR,
a widely used local solver for BM (\cref{sec:local_solvers}) whose performance is highly dependent on the
conditioning of the trust-region subproblem that is solved at each iteration.
% For example, the performance of RTR is highly dependent on the
% conditioning of the trust-region subproblem that is solved at each iteration
% (see \cref{sec:local_solvers}).

Intuitively, preconditioning in the case of RTR attempts to
transform the trust-region subproblem's loss-landscape from highly elongated to spherical (i.e.,
isotropic),
allowing for more efficient iterations towards the solution.
In practical implementations, preconditioning is often carried out by transforming the current search direction (e.g., provided by the negative gradient) via a symmetric and positive definite map $P$.
For perception applications, which are often poorly conditioned,
we have observed that a suitable preconditioner is often indispensable for RTR
% to obtain a satisfactory level of accuracy within
to obtain an acceptably accurate solution within
the runtime constraints of real-time robotics.
In particular, choosing a {good} preconditioner has often led to several orders
of magnitude improvements in terms of (i) number of tCG iterations to converge to a suitable
trust-region subproblem solution, and (ii) overall runtime.

% In the case of ill-conditioned problems, a suitable preconditioner is often the difference between successful
% and unsuccessful optimization.
% \textit{Our experience:} In the applications we have faced,
% a preconditioner has often been necessary to converge to any solution at all.
% Furthermore, choosing a \textit{good} preconditioner has often led several orders
% of magnitude improvements in terms of (i) number of iterations to converge to a suitable
% trust-region subproblem solution, and (ii) overall runtime.
% For us, preconditioning has been the
% difference between real-time robotics and intractable problems.

There is a rich literature and theory behind the construction and analysis of
preconditioners (e.g., \cite[Ch. 10]{saad2003iterative} and
\cite[Ch. 10.2.7]{Golub96book}), which we do not delve into here.  We
instead
outline relevant considerations and suggest a generally successful
preconditioner for the RTR algorithm applied to BM problems.

\textit{Important considerations:} There are roughly three
aspects in which a preconditioner affects the runtime of an optimization
algorithm: (i) the cost of calculating the preconditioner, (ii) the cost
of applying the preconditioner, and (iii) the savings in the number of iterations
required to converge. Ideally a preconditioner is computed once and reused
across many iterations, amortizing the cost of computation. In general,
there is no one-size-fits-all preconditioner, and the choice of preconditioner
depends on the problem structure.

\textit{Preconditioner for RTR:}
For RTR, an ideal preconditioner approximates the inverse of the
Riemannian Hessian of the cost function. The Riemannian Hessian depends on the
point on the manifold at which it is evaluated, and thus typically changes at each
iteration. Importantly, the Riemannian Hessian is closely
related to the Euclidean Hessian \cite[Ch. 5]{boumal2023introduction}.
Furthermore, for the BM formulation we presented
\eqref{eq:bm-problem}, the Euclidean Hessian is exactly the data matrix $Q$ and is
therefore constant. As a result,
$Q^{-1}$ appears as a natural
preconditioner candidate, for it is closely related to the Riemannian Hessian
and can be computed once and used repeatedly. However, preconditioners must be
positive definite. Fortunately,
in the problems we have encountered, $Q$ has been positive
semidefinite and thus becomes positive definite with a small regularization
$Q + \mu I$.

As this would suggest, in the problems we have seen,
the inverse of the regularized data matrix $P = (Q + \mu I)^{-1}$ has been a
successful preconditioner for the RTR trust-region subproblem.
The regularization term $\mu \in \R$ is typically
chosen to keep $P$'s condition number below $10^6$. Instead of
directly computing the inverse, a Cholesky factorization $R^\top R = Q + \mu I$
is computed (with a sparsity-promoting ordering). This allows the
preconditioner to be applied with greater numerical stability and efficiency via
forward- and back-substitution with the Cholesky factor $R$.

\textbf{Sparsity.}
The data and constraint matrices of many problems are often sparse (many
elements are zero). For large-scale problems, exploiting this sparsity can lead
to significant computational savings. Oftentimes, exploiting sparsity involves
using specific data structures, software libraries, and algorithms that are
designed to handle sparse matrices efficiently. Sparsity reduces memory
footprint, improves computational efficiency, and can lead to more numerically
stable procedures.
In distributed settings, sparsity promotes communication efficiency.
Most state-of-the-art distributed optimization methods (e.g., those in
\cref{sec:local_solvers}) seek to preserve and leverage sparsity, so that robots
only exchange information over a small number of variables that couple
together robots' local factor graphs.

\textbf{Existing solvers.}
As previously mentioned, there are a number of general-purpose Riemannian
optimization libraries that have been used for certifiable perception. The
Manopt family \cite{boumal2014manopt,townsend2016pymanopt,bergmann2022manoptjl}
spans MATLAB, Python, and Julia. In C++, there is ROPTLIB
\cite{huang2018roptlib}, GTSAM \cite{gtsam}, and the Optimization library by
\citet{rosenOptimizationLib}. Additionally, while, to our knowledge, Ceres
\cite{ceres-solver} and g2o \cite{g2o} have not been used for certifiable
perception, they could also be used to solve BM problems in the intrinsic form.
In the extrinsic setting, to the best of our knowledge, no standard BM solvers
exist, although practitioners have written custom interfaces to more general
optimization libraries (e.g., \citet{rosen2021scalable} used IPOPT
\cite{wachter2006implementation}).

% : manopt, pymanopt, Dave's, ROPTLIB, \yulunnote{should GTSAM/ceres/g2o be here?}
% \item convergence
% \item apparent benefit of compact feasible sets
% Are conditioning / convergence / convergence rate relevant here? \alannote{I made a section "practical details" where I think these little details could fall into}

%% file: sections/open_directions.tex
\section{Open directions}

BM, and certifiable perception more broadly, has shown great promise in
advancing robotic capabilities. We foresee several exciting frontiers for future
research along these lines. While we focus on BM, aspects of these directions
are also broadly relevant to semidefinite optimization, certifiable perception,
and general optimization.

\textbf{Tools to improve accessibility.} While BM has been applied to a variety of
problems, each new problem requires a bespoke formulation. Obtaining these
formulations typically necessitates substantial algebraic manipulation and an
in-depth understanding of the underlying theory behind BM and certifiable
perception. This represents a significant barrier to entry for practitioners. It
is unknown whether methodologies could be developed to automatically derive
useful QCQP formulations (i.e., that satisfy the LICQ and possess tight SDP
relaxations). Such a tool could possibly leverage the growing catalogue of
successful formulations or develop an approach for finding QCQP formulations
that approximate a given problem. 
Additionally, it may be possible that no such useful QCQP formulations could be
found for a given problem; an impactful tool in this case could assist a user in
determining if such a formulation is likely to be found.
Advances in this direction would greatly
benefit from deep understanding of structural relationships between QCQPs and
their SDP relaxations. 

% In the same way that e.g., GTSAM \cite{gtsam} allows for
% rapid prototyping of optimization problems, an impactful direction is to
% automate the process of deriving BM formulations or otherwise determining if
% such a formulation is likely to be successful.
% we envision a future where BM formulations
% can be derived without requiring substantial algebraic exploration.

\textbf{Robust costs and outlier rejection.} Outlier rejection (via robust cost
functions) has been explored in certifiable perception \cite{yang2020neurips},
however every formulation to date has had to introduce redundant constraints
and, as a result, violate the LICQ. Without the LICQ these formulations have not
been able to leverage the computational benefits of BM. It is an open question
whether it is possible to formulate robust cost functions that preserve the
LICQ.

\textbf{Distributed optimization.} In centralized settings, practitioners
have largely converged to a set of well-implemented trust-region algorithms
(such as RTR and Levenberg-Marquardt).
In the distributed setup, despite the initial progress discussed in earlier sections of this paper,
there is still no such consensus and additional research is required especially when considering limited
communication (either due to bandwidth restrictions or privacy concerns).
% there is no such consensus in the
% distributed setting, which incurs unique challenges when considering limited
% communication (either due to bandwidth restrictions or privacy concerns).
We believe this opens the possibility of contributions in: (i) designing
distributed algorithms for optimization and certification, (ii) analyzing the
convergence of these algorithms, and (iii) developing standard software
tools.

\comment{
\begin{itemize}
    \item Convergence rate in distributed
    \item What do you do when LICQ isn't satisfied (besides accept slow results)?
    \item incredibly bespoke - hard to find generalizations (what does GTSAM for this look like?)
    \item Possibility of introducing robustness/outlier rejection that preserves the LICQ
    \item problems that don't fit the mold of the Stiefel manifold
\end{itemize}
}

%% file: sections/conclusion.tex
\section{Conclusion}

We have presented an introductory overview of the Burer-Monteiro method (BM) for
certifiable perception problems. We discussed key theoretical properties,
outlined important theoretical requirements that are not typically discussed in
the literature, and provided practical considerations for applying BM to
perception problems. We also discussed open directions for future research in
this area.

We believe that BM will play an
important role in the future of robotics and perception.  However, we also
believe that BM is not a one-size-fits-all solution. BM should be used when
both of the following conditions are met: (i) tight SDP relaxations can be
constructed via Shor's relaxation and (ii) the QCQP formulation globally satisfies
the LICQ.
However, many perception problems will likely not be able to satisfy
these conditions.
Indeed, better understanding the boundary of BM's applicability and determining
when alternative approaches (e.g.,
\cite{duembgen2024globally,barfoot2024certifiably}) are important
questions that demand more practical insights and theoretical investigations.

%% file: sections/acknowledgements.tex
\section{Acknowledgements}
% YULUN
% This work was supported in part by ARL
% DCIST under Cooperative Agreement Number W911NF-17-20181, and in part by ONR
% under BRC Award N000141712072.”

% % JLeonard
% ONR grants N00014-23-12164, and N00014-19-1-2571 (Neuroautonomy MURI), the
% MIT Portugal Program, and the MIT Lincoln Laboratory Autonomous
% Systems Line which is funded by the Under Secretary of Defense for
% Research and Engineering through Air Force Contract No. FA8702-15-D-0001.

This work was supported in part by ARL DCIST under Cooperative Agreement Number
W911NF-17-20181, by ONR under BRC Award N000141712072, by ONR grants
N00014-23-12164, and N00014-19-1-2571 (Neuroautonomy MURI), the MIT Portugal
Program, and the MIT Lincoln Laboratory Autonomous Systems Line which is funded
by the Under Secretary of Defense for Research and Engineering through Air Force
Contract No. FA8702-15-D-0001, and
MIT Lincoln Laboratory's Certifiable Semialgebraic Optimization program under contract FA8702-15-D-0001.

%% file: main.bbl
\begin{thebibliography}{46}
\providecommand{\natexlab}[1]{#1}
\providecommand{\url}[1]{\texttt{#1}}
\expandafter\ifx\csname urlstyle\endcsname\relax
  \providecommand{\doi}[1]{doi: #1}\else
  \providecommand{\doi}{doi: \begingroup \urlstyle{rm}\Url}\fi

\bibitem[Burer and Monteiro(2003)]{burer2003nonlinear}
Samuel Burer and Renato~D.C. Monteiro.
\newblock A nonlinear programming algorithm for solving semidefinite programs via low-rank factorization.
\newblock \emph{Mathematical programming}, 95\penalty0 (2):\penalty0 329--357, 2003.

\bibitem[Rosen(2021)]{rosen2021scalable}
David~M. Rosen.
\newblock Scalable low-rank semidefinite programming for certifiably correct machine perception.
\newblock In \emph{Algorithmic Foundations of Robotics XIV: Proceedings of the Fourteenth Workshop on the Algorithmic Foundations of Robotics 14}, pages 551--566. Springer, 2021.

\bibitem[Tian et~al.(2021)Tian, Khosoussi, Rosen, and How]{tian2021distributed}
Yulun Tian, Kasra Khosoussi, David~M. Rosen, and Jonathan~P How.
\newblock Distributed certifiably correct pose-graph optimization.
\newblock \emph{IEEE Transactions on Robotics}, 37\penalty0 (6):\penalty0 2137--2156, 2021.

\bibitem[Papalia et~al.(2024)Papalia, Fishberg, O'Neill, How, Rosen, and Leonard]{papalia24tro}
Alan Papalia, Andrew Fishberg, Brendan~W. O'Neill, Jonathan~P. How, David~M. Rosen, and John~J. Leonard.
\newblock Certifiably correct range-aided {SLAM}.
\newblock \emph{IEEE Transactions on Robotics}, pages 1--20, 2024.
\newblock \doi{10.1109/TRO.2024.3454430}.

\bibitem[Shor(1987)]{shor1987quadratic}
Naum~Z. Shor.
\newblock Quadratic optimization problems.
\newblock \emph{Soviet Journal of Computer and Systems Sciences}, 25:\penalty0 1--11, 1987.

\bibitem[Vandenberghe and Boyd(1996)]{vandenberghe1996semidefinite}
Lieven Vandenberghe and Stephen Boyd.
\newblock Semidefinite programming.
\newblock \emph{SIAM review}, 38\penalty0 (1):\penalty0 49--95, 1996.

\bibitem[Rosen et~al.(2019)Rosen, Carlone, Bandeira, and Leonard]{rosen2019se}
David~M. Rosen, Luca Carlone, Afonso~S. Bandeira, and John~J. Leonard.
\newblock {SE-Sync}: A certifiably correct algorithm for synchronization over the special {Euclidean} group.
\newblock \emph{The International Journal of Robotics Research}, 38\penalty0 (2-3):\penalty0 95--125, 2019.

\bibitem[Cifuentes et~al.(2022)Cifuentes, Agarwal, Parrilo, and Thomas]{cifuentes2022local}
Diego Cifuentes, Sameer Agarwal, Pablo~A Parrilo, and Rekha~R Thomas.
\newblock On the local stability of semidefinite relaxations.
\newblock \emph{Mathematical Programming}, pages 1--35, 2022.

\bibitem[Majumdar et~al.(2020)Majumdar, Hall, and Ahmadi]{majumdar2020recent}
Anirudha Majumdar, Georgina Hall, and Amir~Ali Ahmadi.
\newblock Recent scalability improvements for semidefinite programming with applications in machine learning, control, and robotics.
\newblock \emph{Annual Review of Control, Robotics, and Autonomous Systems}, 3:\penalty0 331--360, 2020.

\bibitem[Boumal(2015)]{boumal2015riemannian}
Nicolas Boumal.
\newblock A {Riemannian} low-rank method for optimization over semidefinite matrices with block-diagonal constraints.
\newblock \emph{arXiv preprint arXiv:1506.00575}, 2015.

\bibitem[W{\"a}chter and Biegler(2006)]{wachter2006implementation}
Andreas W{\"a}chter and Lorenz~T. Biegler.
\newblock On the implementation of an interior-point filter line-search algorithm for large-scale nonlinear programming.
\newblock \emph{Mathematical programming}, 106:\penalty0 25--57, 2006.

\bibitem[Briales and Gonzalez-Jimenez(2017)]{briales2017cartan}
Jesus Briales and Javier Gonzalez-Jimenez.
\newblock {Cartan-Sync}: Fast and global {SE}(d)-synchronization.
\newblock \emph{IEEE Robotics and Automation Letters}, 2\penalty0 (4):\penalty0 2127--2134, 2017.

\bibitem[Dellaert et~al.(2020)Dellaert, Rosen, Wu, Mahony, and Carlone]{dellaert2020shonan}
Frank Dellaert, David~M. Rosen, Jing Wu, Robert Mahony, and Luca Carlone.
\newblock Shonan rotation averaging: Global optimality by surfing ${SO}(p)^n$.
\newblock In \emph{Computer Vision--ECCV 2020: 16th European Conference, Glasgow, UK, August 23--28, 2020, Proceedings, Part VI 16}, pages 292--308. Springer, 2020.

\bibitem[Absil et~al.(2008)Absil, Mahony, and Sepulchre]{absil2008optimization}
P-A Absil, Robert Mahony, and Rodolphe Sepulchre.
\newblock \emph{Optimization algorithms on matrix manifolds}.
\newblock Princeton University Press, 2008.

\bibitem[Boumal(2023)]{boumal2023introduction}
Nicolas Boumal.
\newblock \emph{An introduction to optimization on smooth manifolds}.
\newblock Cambridge University Press, 2023.

\bibitem[Absil et~al.(2007)Absil, Baker, and Gallivan]{absil2007trust}
P.-A. Absil, Christopher~G. Baker, and Kyle~A. Gallivan.
\newblock Trust-region methods on {Riemannian} manifolds.
\newblock \emph{Foundations of Computational Mathematics}, 7:\penalty0 303--330, 2007.

\bibitem[{Frank Dellaert et al.}(2019)]{gtsam}
{Frank Dellaert et al.}
\newblock {Georgia Tech Smoothing And Mapping (GTSAM)}.
\newblock \url{https://gtsam.org/}, 2019.

\bibitem[Tian et~al.(2020)Tian, Koppel, Bedi, and How]{tian2020asynchronous}
Yulun Tian, Alec Koppel, Amrit~Singh Bedi, and Jonathan~P. How.
\newblock Asynchronous and parallel distributed pose graph optimization.
\newblock \emph{IEEE Robotics and Automation Letters}, 5\penalty0 (4):\penalty0 5819--5826, 2020.

\bibitem[Peng and Vidal(2023)]{peng2023block}
Liangzu Peng and Rene Vidal.
\newblock Block coordinate descent on smooth manifolds: Convergence theory and twenty-one examples.
\newblock In \emph{Conference on Parsimony and Learning (Recent Spotlight Track)}, 2023.

\bibitem[Tron and Vidal(2009)]{tron2009distributed}
Roberto Tron and Ren{\'e} Vidal.
\newblock Distributed image-based {3-D} localization of camera sensor networks.
\newblock In \emph{Proceedings of the 48h IEEE Conference on Decision and Control (CDC) held jointly with 2009 28th Chinese Control Conference}, pages 901--908. IEEE, 2009.

\bibitem[Tron and Vidal(2014)]{tron2014distributed}
Roberto Tron and Ren{\'e} Vidal.
\newblock Distributed {3-D} localization of camera sensor networks from {2-D} image measurements.
\newblock \emph{IEEE Transactions on Automatic Control}, 59\penalty0 (12):\penalty0 3325--3340, 2014.

\bibitem[Asgharivaskasi et~al.(2024)Asgharivaskasi, Girke, and Atanasov]{asgharivaskasi2024riemannian}
Arash Asgharivaskasi, Fritz Girke, and Nikolay Atanasov.
\newblock Riemannian optimization for active mapping with robot teams.
\newblock \emph{arXiv preprint arXiv:2404.18321}, 2024.

\bibitem[Xu et~al.(2022)Xu, Liu, Chen, and Shen]{xu2022d}
Hao Xu, Peize Liu, Xinyi Chen, and Shaojie Shen.
\newblock ${D}^{2}${SLAM}: Decentralized and distributed collaborative visual-inertial {SLAM} system for aerial swarm.
\newblock \emph{arXiv preprint arXiv:2211.01538}, 2022.

\bibitem[B{\"a}nninger et~al.(2023)B{\"a}nninger, Alzugaray, Karrer, and Chli]{banninger2023cross}
Philipp B{\"a}nninger, Ignacio Alzugaray, Marco Karrer, and Margarita Chli.
\newblock Cross-agent relocalization for decentralized collaborative {SLAM}.
\newblock In \emph{2023 IEEE International Conference on Robotics and Automation (ICRA)}, pages 5551--5557. IEEE, 2023.

\bibitem[McGann et~al.(2024)McGann, Lassak, and Kaess]{mcgann2023asynchronous}
Daniel McGann, Kyle Lassak, and Michael Kaess.
\newblock Asynchronous distributed smoothing and mapping via on-manifold consensus {ADMM}.
\newblock \emph{2024 IEEE International Conference on Robotics and Automation (ICRA)}, pages 4577--4583, 2024.

\bibitem[Fan and Murphey(2023)]{fan2023majorization}
Taosha Fan and Todd~D. Murphey.
\newblock Majorization minimization methods for distributed pose graph optimization.
\newblock \emph{IEEE Transactions on Robotics}, 2023.

\bibitem[Boyd and Vandenberghe(2004)]{boyd2004convex}
Stephen~P. Boyd and Lieven Vandenberghe.
\newblock \emph{Convex Optimization}.
\newblock Cambridge University Press, 2004.

\bibitem[Boumal et~al.(2014)Boumal, Mishra, Absil, and Sepulchre]{boumal2014manopt}
Nicolas Boumal, Bamdev Mishra, P-A Absil, and Rodolphe Sepulchre.
\newblock Manopt, a {Matlab} toolbox for optimization on manifolds.
\newblock \emph{The Journal of Machine Learning Research}, 15\penalty0 (1):\penalty0 1455--1459, 2014.

\bibitem[Nocedal and Wright(2006)]{nocedal2006book}
Jorge Nocedal and Stephen~J. Wright.
\newblock \emph{Numerical optimization}.
\newblock Springer, 2006.

\bibitem[Wachsmuth(2013)]{wachsmuth13licq}
Gerd Wachsmuth.
\newblock On {LICQ} and the uniqueness of {L}agrange multipliers.
\newblock \emph{Operations Research Letters}, 41\penalty0 (1):\penalty0 78--80, 2013.

\bibitem[Fan et~al.(2020)Fan, Wang, Rubenstein, and Murphey]{fan2020cpl}
Taosha Fan, Hanlin Wang, Michael Rubenstein, and Todd Murphey.
\newblock {CPL}-{SLAM}: Efficient and certifiably correct planar graph-based {SLAM} using the complex number representation.
\newblock \emph{IEEE Transactions on Robotics}, 36\penalty0 (6):\penalty0 1719--1737, 2020.

\bibitem[Holmes and Barfoot(2023)]{holmes2023efficient}
Connor Holmes and Timothy~D. Barfoot.
\newblock An efficient global optimality certificate for landmark-based {SLAM}.
\newblock \emph{IEEE Robotics and Automation Letters}, 8\penalty0 (3):\penalty0 1539--1546, 2023.

\bibitem[Halsted and Schwager(2022)]{halsted2022riemannian}
Trevor Halsted and Mac Schwager.
\newblock The {Riemannian} elevator for certifiable distance-based localization.
\newblock \emph{Preprint}, 2022.

\bibitem[Karimian and Tron(2023)]{karimian2023essential}
Arman Karimian and Roberto Tron.
\newblock Essential matrix estimation using convex relaxations in orthogonal space.
\newblock In \emph{Proceedings of the IEEE/CVF International Conference on Computer Vision}, pages 17142--17152, 2023.

\bibitem[Hu and Carlone(2019)]{hu2019accelerated}
Siyi Hu and Luca Carlone.
\newblock Accelerated inference in {Markov} random fields via smooth {Riemannian} optimization.
\newblock \emph{IEEE Robotics and Automation Letters}, 4\penalty0 (2):\penalty0 1295--1302, 2019.

\bibitem[Saad(2003)]{saad2003iterative}
Yousef Saad.
\newblock \emph{Iterative methods for sparse linear systems}.
\newblock SIAM, 2003.

\bibitem[Golub and Loan(1996)]{Golub96book}
Gene Golub and Charles~Van Loan.
\newblock \emph{Matrix Computations}.
\newblock Johns Hopkins University Press, Baltimore, MD, 3rd edition, 1996.

\bibitem[Townsend et~al.(2016)Townsend, Koep, and Weichwald]{townsend2016pymanopt}
James Townsend, Niklas Koep, and Sebastian Weichwald.
\newblock Pymanopt: A python toolbox for optimization on manifolds using automatic differentiation.
\newblock \emph{Journal of Machine Learning Research}, 17\penalty0 (137):\penalty0 1--5, 2016.

\bibitem[Bergmann(2022)]{bergmann2022manoptjl}
Ronny Bergmann.
\newblock Manopt.jl: Optimization on manifolds in {J}ulia.
\newblock \emph{Journal of Open Source Software}, 7\penalty0 (70):\penalty0 3866, 2022.
\newblock \doi{10.21105/joss.03866}.

\bibitem[Huang et~al.(2018)Huang, Absil, Gallivan, and Hand]{huang2018roptlib}
Wen Huang, P.-A. Absil, Kyle~A. Gallivan, and Paul Hand.
\newblock {ROPTLIB}: an object-oriented {C++} library for optimization on {Riemannian} manifolds.
\newblock \emph{ACM Transactions on Mathematical Software (TOMS)}, 44\penalty0 (4):\penalty0 1--21, 2018.

\bibitem[Rosen(2022)]{rosenOptimizationLib}
David~M. Rosen.
\newblock {Optimization}, 7 2022.
\newblock URL \url{https://github.com/david-m-rosen/Optimization}.

\bibitem[{Sameer Agarwal et al.}(2024)]{ceres-solver}
{Sameer Agarwal et al.}
\newblock Ceres solver.
\newblock \url{http://ceres-solver.org}, 2024.

\bibitem[Grisetti et~al.(2011)Grisetti, K{\"u}mmerle, Strasdat, and Konolige]{g2o}
Giorgio Grisetti, Rainer K{\"u}mmerle, Hauke Strasdat, and Kurt Konolige.
\newblock g2o: A general framework for (hyper) graph optimization.
\newblock In \emph{Proceedings of the IEEE international conference on robotics and automation (ICRA), Shanghai, China}, pages 9--13, 2011.

\bibitem[Yang and Carlone(2020)]{yang2020neurips}
Heng Yang and Luca Carlone.
\newblock One ring to rule them all: Certifiably robust geometric perception with outliers.
\newblock In H.~Larochelle, M.~Ranzato, R.~Hadsell, M.F. Balcan, and H.~Lin, editors, \emph{Advances in Neural Information Processing Systems}, volume~33, pages 18846--18859. Curran Associates, Inc., 2020.
\newblock URL \url{https://proceedings.neurips.cc/paper_files/paper/2020/file/da6ea77475918a3d83c7e49223d453cc-Paper.pdf}.

\bibitem[Dümbgen et~al.(2024)Dümbgen, Holmes, Agro, and Barfoot]{duembgen2024globally}
Frederike Dümbgen, Connor Holmes, Ben Agro, and Timothy~D. Barfoot.
\newblock Toward globally optimal state estimation using automatically tightened semidefinite relaxations.
\newblock 2024.

\bibitem[Barfoot et~al.(2024)Barfoot, Holmes, and Dümbgen]{barfoot2024certifiably}
Timothy~D Barfoot, Connor Holmes, and Frederike Dümbgen.
\newblock Certifiably optimal rotation and pose estimation based on the {Cayley} map.
\newblock 2024.

\end{thebibliography}
